\documentclass[a4paper, 10pt, conference]{ieeeconf}   
\usepackage{FG2019}
\usepackage{float}
\usepackage{subcaption}
\usepackage{multirow}
\usepackage{amsmath,graphicx,amssymb}
\usepackage{pifont}
\usepackage[nocompress]{cite}
\usepackage[bookmarks=false]{hyperref}
\usepackage{soul} 
\newcommand{\cmark}{\ding{51}}%
\newcommand{\xmark}{\ding{55}}%

\newcommand{\christy}[1]{\textcolor{black}{#1}}
\newcommand{\ice}[1]{\textcolor{black}{#1}}
\FGfinalcopy 

\def\FGPaperID{3} 

\title{\LARGE \bf
Shallow Triple Stream Three-dimensional CNN (STSTNet) for Micro-expression Recognition}

\author{\parbox{16cm}{\centering
    {\large Sze-Teng Liong$^1$, Y.S. Gan$^2$, John See$^3$, Huai-Qian Khor$^3$, Yen-Chang Huang$^4$}\\
    {\normalsize
    $^1$Department of Electronic Engineering, Feng Chia University, Taichung 40724, Taiwan R.O.C.\\
    $^2$ Department of Info. Management, National Taipei University of Nursing and Health Sciences, Taiwan R.O.C.\\
    $^3$ Faculty of Computing and Informatics, Multimedia University, 63100 Cyberjaya, Malaysia\\
    $^4$  School of Mathematics and Statistics, Xinyang Normal University, Henan, China}}
}

\begin{document}

\ifFGfinal
\thispagestyle{empty}
\pagestyle{empty}
\else
\author{Anonymous FG 2019 submission\\ Paper ID \FGPaperID \\}
\pagestyle{plain}
\fi
\maketitle

\begin{abstract}
In the recent year, state-of-the-art for facial micro-expression recognition have been significantly advanced by deep neural networks.
The robustness of deep learning has yielded promising performance beyond that of traditional handcrafted approaches.
 Most works in literature emphasized on increasing the depth of networks and employing highly complex objective functions to learn more features.
\christy{In this paper, we design a Shallow Triple Stream Three-dimensional CNN (STSTNet) that is computationally light whilst capable of extracting discriminative high level features and details of micro-expressions. }
The network 
learns
from three optical flow features (i.e., optical strain, horizontal and vertical optical flow fields) computed based on the onset and apex frames of each video.
Our experimental results demonstrate the effectiveness of the proposed STSTNet, which obtained an unweighted average recall rate of 0.7605 and unweighted F1-score of 0.7353 on the composite database consisting of 442 samples from the SMIC, CASME II and SAMM databases.  

\end{abstract}

\section{Introduction}
\christy{Facial expressions are a form of nonverbal communication created by facial muscle contractions during emotional states.}
Different muscular movements and patterns eventually reflect different types of emotions.
\christy{However, the expressions portray on the face may not accurately imply one's emotion state as it can be faked easily. }

Among several types of nonverbal communications (i.e., facial expression, vocal intonation and body posture), micro-expression (ME) is discovered to be the likeliest to reveal one’s deepest emotions~\cite{ekman2009telling}. 
Since the ME is stimulated involuntarily, it allows the \ice{competence} in exposing one’s concealed genuine feelings without deliberate control.
In contrast to facial macro-expressions, which normally lasts between 0.75s to 2s, micro-expression usually \christy{occurs} in less duration (0.04s to 0.2s) and lower intensity~\cite{ekman1971constants}.

In recent years, there has been a \ice{ growing} interest in \ice{incorporating computer vision techniques in} automated ME recognition systems. 
The state-of-the-art approaches for ME recognition (based on original protocol) have obtained accuracy levels less than 70\%~\cite{allaert2017consistent, li2018towards}, \ice{
though tested on the datasets constructed in constrained laboratory environment.} 
In contrast, normal (macro-) expression recognition systems can exhibit almost perfect recognition accuracy~\cite{kharat2009emotion, rivera2013local}. 
\ice{Meanwhile, }\christy{most of the ME videos are captured using high frame rate cameras (i.e. $>$100$fps$)} and resulting in a lot of redundant frames. \ice{Hence,} it is essential to eliminate the overload of unnecessary facial information while highlighting important characteristics and cues of ME movements. 
Temporal Interpolation Method (TIM)~\cite{zhou2012image} is one of the techniques used in ME systems to address the problem of different video lengths~\cite{li2018towards, wang2015RPCA, pfister2011}.
It normalizes the length of all image sequences to a certain fixed length, either through downsampling or upsampling. 
TIM was adopted by the original ME databases 
to standardize the frame length before feature extraction.
Moving along these lines, Liong et al.~\cite{liong2018less,liong2018off} proposed to identify the ME category by using only information from a single apex frame (i.e., frame with highest emotion intensity). They demonstrated that it is sufficient to encode \ice{ME} features by utilizing the apex and a neutral reference frame (typically the onset). 

For feature extraction, \ice{numerous} researchers proposed algorithms based on Local Binary Pattern (LBP)~\cite{ojala1996}, such as  
Local Binary Pattern on Three-Orthogonal Planes (LBP-TOP)~\cite{zhao2007}, Local Binary Pattern with Six Intersection Points (LBP-SIP)~\cite{wang2015lbp}, Local Binary Pattern with Mean Orthogonal Planes (LBP-MOP)~\cite{wang2015efficient} and Spatiotemporal Completed Local Quantization Patterns (STCLQP)~\cite{huang2016spontaneous}.
LBP is a texture-based feature extraction method \ice{with} characteristics of good discrimination ability, compact representation and low computational complexity. 
\ice{Meanwhile, some works favor optical flow features} that estimate the \ice{frame-level motions} based on the change in brightness intensities between frames, which is capable \ice{of capturing subtle facial movements}. Optical flow-based approaches include Optical Strain Feature (OSF)~\cite{liong2014optical}, Optical Strain Weight (OSW)~\cite{liong2014subtle}, Fuzzy Histogram of Oriented Optical Flow (FHOOF)~\cite{happy2017fuzzy}, Main Directional Mean optical flow (MDMO)~\cite{liu2016mdmo} and most recently, Bi-Weighted Oriented Optical Flow (Bi-WOOF)~\cite{liong2018less}.

One of the earliest ME works that adopted convolutional neural network (CNN) is one by Patel et al.~\cite{peter2016}.
However, their method \ice{fared poorer than} many conventional handcrafted descriptors \ice{
due to the possibility of model overfitting}. 
On the other hand, a recent work by Li et al.~\cite{li2018can} \ice{
finetuned a VGG-Face model with ME apex frames and achieved up to $\sim$63\% in accuracy but at the expense of enormous trainable parameters (i.e., 138 Million).}
\ice{
Likewise}, Wang et al.~\cite{wang2018micro} adopted a CNN and Long Short-Term Memory (LSTM) architecture to learn the spatial-temporal information for each image sequence, \ice{
which also comes with huge number of parameters (i.e. 80 Million)}.
Prior to passing the image frames into the model, TIM is applied to each video sequence, 
to standardize the frame length to either 32 or 64.
Besides, a three-stream CNN network is proposed by Li et al.~\cite{li2018micro} where each stream takes in a different type of data: grayscale, horizontal and vertical optical flow fields.
\ice{Empirically, it performed as good as some recent methods ($\sim$60\%)~\cite{huang2016spontaneous, liong2018less} for CASME~II but ineffective in SMIC with a mere $\sim$55\% accuracy.}

To the best of our knowledge,~\cite{liong2018off} is the first work \ice{that} \christy{performs} cross-dataset validation on three distinct databases (i.e., CASME II~\cite{casme2}, SMIC~\cite{smic}, SAMM~\cite{samm}).
Succinctly, they \ice{proposed} a three-step framework:
1) Apex frame acquisition from each video;
2) Computation for optical flow guided features (i.e., horizontal and vertical optical flow images) from the apex and onset frames;
3) Feature learning and fusion 
using a neural network (coined as `OFF-ApexNet').
Hence, motivated by~\cite{liong2018off}, this paper aims to improve the recognition performance by further simplification of the neural network \ice{while preserving} sufficient capacity to learn the real structure of the ME details. 
The main contributions of this paper are as follows:
\begin{enumerate}
\item Proposal of a small and shallow 3-D convolutional neural network whilst preserving the effectiveness in generating rich and discriminative feature representation.
\item Feature extraction from three optical flow information (i.e., optical strain, horizontal and vertical optical flow).
\item Re-implementation of several state-of-the-art methods and baseline CNN architectures, and providing \ice{quantitative experimental analyses.} 

\end{enumerate}

\section{Proposed Method}

\begin{figure}[thpb]
\centering
\includegraphics[width=1\linewidth]{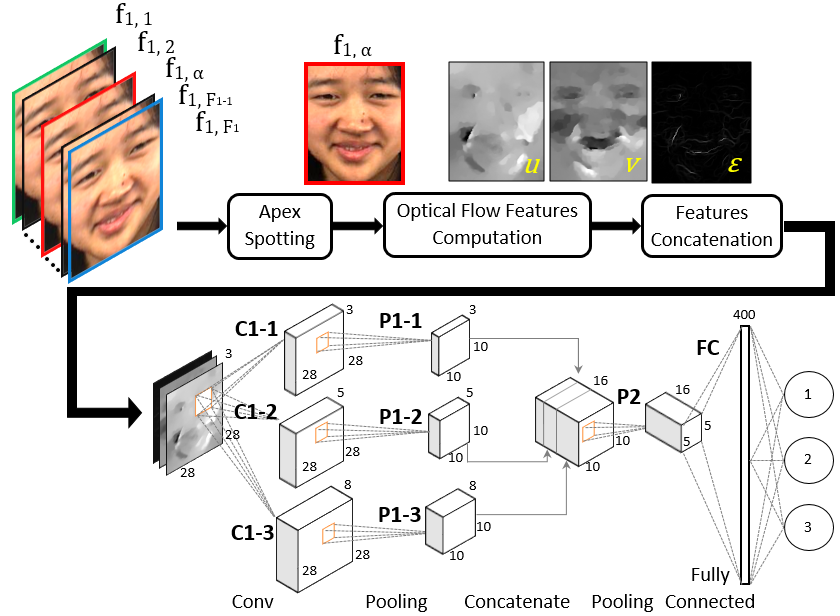}
\caption{Flow diagram of proposed STSTNet approach  
}
\label{fig:flow}
\end{figure}

While many architectures proposed in the literature relied on increasing the number of neurons or increasing the number of layers to allow the network to learn more complex functions, this paper presents a shallow neural network architecture that comprises \ice{of} two learnable layers.
Similar to~\cite{liong2018off}, the proposed micro-expression recognition scheme consists of three main steps, namely: apex frame spotting, optical flow features computation and feature learning with CNN. 
The overview of the recognition approach is illustrated in Fig.~\ref{fig:flow}.

\subsection{Apex frame spotting}
Firstly, the apex frame spotting stage is to identify the frame that contains the highest intensity of ME in a video sequence.
Since SMIC database does not provide the ground-truth apex frame, we employ the D\&C-RoIs~\cite{liong2015spotting} approach to obtain the apex frame index.
D\&C-RoIs has been utilized by several recent ME works~\cite{liong2018less,liong2018off, liong2017phase, gan2018bi} as it facilitates in producing reasonably good performance for the purpose of ME recognition.
With the LBP descriptor as feature choice, the method first computes the correlation between the first frame and the rest of the frames:
\begin{equation}
    d = \frac{\sum_{i=1}^{B}h_{1i}\times h_{2i}}{\sqrt{\sum_{i=1}^{B}h_{1i}^{2}\times \sum_{i=1}^{B}h_{2i}^{2} }}
    \label{eq:lbp-corr}
\end{equation}
where B is the number of bins in histograms $h_1$ (first frame), and
$h_2$ ($N-1$ other frames). The rate of difference ($1-d$) of the LBP features are then compared among the three ROIs and the ROI with the highest rate of difference is selected. 
Finally, a divide-and-conquer strategy is applied to \christy{search} for the frame \ice{with} maximum facial muscle changes.

For clarity, we define some notations for ease of explanations. 
Within a collection of video data, the $i$-th ME video sequence consists of $F_i$ number of frames:
\begin{equation}
s_{i} = \{f_{i,j} | i=1,\dots,n; j=1,\dots ,F_{i}\},
\end{equation}
Note that each video contains only one onset (starting) frame $f_{i,1}$, one offset (ending) frame $f_{i,F_i}$ and a single apex frame $f_{i,\alpha} \in [f_{i,1}, f_{i,F_i}]$. 
Note that the apex frame for SMIC 
is obtained via the D\&C-RoIs spotting approach, while the other two databases have already provided apex frame annotations. 

\subsection{Optical flow guided features}

Next, we compute optical flow guided features using the onset and apex frames.
The optical flow field that is computed from these two frames can be formulated as a tuple:
\begin{equation}
O_i = \{(u(x,y), v(x,y)) | x = 1, 2, ... , X, y = 1, ... , Y\},
\end{equation}
where $X$ and $Y$ denote the width and height of the frame $f_{i,j}$, respectively, while
$u(x, y)$ and $v(x, y)$ represent the horizontal and vertical components of $O_i$, respectively.
Another optical flow derivative, known as optical strain, is capable of approximating the intensity of facial deformation and it can be defined as: 
\begin{equation} \label{eq:tensor}
\varepsilon = \frac{1}{2} [\nabla \bf u + (\nabla \bf u)^{\it T} ],
\end{equation}
where {\bf u} = $[u,v]^T$ is the displacement vector. It can also be re-written as a Hessian matrix:
\begin{equation}
\varepsilon = \begin{bmatrix}
      		\varepsilon_{xx} = \frac{\partial u}{\partial x} & \varepsilon_{xy} = \frac{1}{2}(\frac{\partial u}{\partial y} + \frac{\partial v}{\partial x}) \\[1em]
      	    \varepsilon_{yx} = \frac{1}{2}(\frac{\partial v}{\partial x} + \frac{\partial u}{\partial y}) & \varepsilon_{yy} = \frac{\partial v}{\partial y}
     		\end{bmatrix},
\end{equation}

\noindent
where the diagonal terms, ($\varepsilon_{xx},\varepsilon_{yy}$), are normal strain components and ($\varepsilon_{xy},\varepsilon_{yx}$) are shear strain components. 
The optical strain magnitude for each pixel can be computed by taking the sum of squares of the normal and shear strain components, such that:
\begin{equation}
\begin{split}
|\varepsilon_{x,y}| 
& = \sqrt{{\frac{\partial u}{\partial x}}^{2} + {\frac{\partial v}{\partial y}}^{2} +\frac{1}{2}{(\frac{\partial u}{\partial x} + \frac{\partial u}{\partial x})}^{2}}.
\end{split}
\label{eq:osm}
\end{equation}

Appending the optical strain to the optical flow field $O$ yields a triple, $\Theta=\{u, v, \epsilon \} \in \mathbb{R}^{3}$. In summary, each video can be derived into the following three optical flow based representations:
\begin{itemize}
\item  $u$  - Horizontal component of the optical flow field $O_i$,
\item $v$ - Vertical component of the optical flow field $O_i$, 
\item $\varepsilon$ - Optical strain
\end{itemize}

\subsection{Shallow triple stream 3D-CNN}

The final step is to further learn optical flow guided features using a new shallow triple stream 3D-CNN. By virtue of being a 3D-CNN, all convolutional layers are in three dimensions. The input to the network is the optical flow cube $\Theta$ described in the previous sub-section.
\christy{Following the suggestion in~\cite{liong2018off}, the 3D input cube will be resampled to 28$\times$28$\times$3.}
Then, the image is passed through three parallel streams, each consists of a convolutional layer (each stream has a different number of kernels, i.e. 3, 5, 8) followed by a \ice{max} pooling layer.
This design supplements the small scale input data \ice{by utilizing} different number of 3$\times$3 kernels on each stream to avoid the problem of underfitting the data. \ice{In addition, the max} pooling operation is used to highlight dominant features while eliminating redundancy.
Next, the outputs are merged channel-wise to form a 3D block of features before applying an additional $2\times 2$ average pooling layer.
Lastly, a 400-node fully connected (FC) layer provides further abstraction before the final softmax layer classifies to one of the three ME composite emotion classes. 


The exact network configuration is shown in Table~\ref{table:CNNsetting}.
For all our experiments, we use a small learning rate of $5\times10^{-5}$ with the maximum number of epochs set to 500.

\begin{table}[thpb] 
\begin{center}
\caption{STSTNet configuration for the convolutional (\textit{C}) layers, pooling (\textit{P}) layers, fully connected (\textit{FC}) layer and output sofmax layer
}
\label{table:CNNsetting}
\begin{tabular}{@{}cccccc@{}}
\hline
Layer
& Filter size
& \# Filters
& Stride
& Padding
& Output size \\

\hline

C1-1
& 3 $\times$ 3 $\times$ 3 
& 3
& [1,1]
& 1
& 28 $\times$ 28 $\times$ 3 \\

C1-2
& 3 $\times$ 3 $\times$ 3 
& 5
& [1,1]
& 1
& 28 $\times$ 28 $\times$ 5 \\

C1-3
& 3 $\times$ 3 $\times$ 3 
& 8
& [1,1]
& 1
& 28 $\times$ 28 $\times$ 8 \\

P1-1
& 3 $\times$ 3 
& -
& [3,3]
& 1
& 10 $\times$ 10 $\times$ 3 \\

P1-2
& 3 $\times$ 3 
& -
& [3,3]
& 1
& 10 $\times$ 10 $\times$ 5 \\

P1-3
& 3 $\times$ 3 
& -
& [3,3]
& 1
& 10 $\times$ 10 $\times$ 8 \\

P2
& 2 $\times$ 2 
& -
& [2,2]
& 0
& 5 $\times$ 5 $\times$ 16 \\

FC 
& -
& -
& -
& -
& 400 $\times$ 1 \\

Softmax
& -
& -
& -
& -
& 3 $\times$ 1 \\
\hline
\end{tabular}
\end{center}
\vspace{-5mm}
\end{table}

\section{Experiment}
\subsection{Databases}

There are three databases commonly used for ME recognition: SMIC~\cite{smic}, CASME II~\cite{casme2} and SAMM~\cite{samm,davison2018objective}. 
The detailed information of these three databases are shown in Table~\ref{table:database}. 
It is observed that the databases are  largely limited on their own, and \ice{have} an imbalanced distribution of samples per emotion.
In the second Facial Micro-expression Grand Challenge, the majority of video samples from these three databases are merged into a composite database by mapping their individual emotion classes into three generic emotion classes: `Positive', `Negative' and `Surprise'. There are a total of 442 samples after merging the databases.  

\begin{table}[thpb]
\begin{center}
\caption{Detailed information of the three merged ME databases}
\label{table:database}
\def\arraystretch{0.6}
\begin{tabular}{p{1.4cm}lccc}
\hline
 \multicolumn{2}{l} {Database} & CASME II & SMIC & SAMM  \\
\hline

\multicolumn{2}{l}{Subjects}
&	24 & 16 & 28 \\
\hline

\multicolumn{2}{l}{Samples}
&	145 & 164 & 133 \\
\hline

\multicolumn{2}{l}{Frame rate (\textit{fps})}
&	200 & 100 & 200 \\
\hline

\multicolumn{2}{l}{
\multirow{3}{*}{\begin{tabular}[c]{@{}l@{}} Cropped image resolution\\  (pixels)\end{tabular}} }
& \multirow{3}{*}{170 $\times$ 140}
& \multirow{3}{*}{170 $\times$ 140}
& \multirow{3}{*}{170 $\times$ 140} \\\\\\
\hline
 
\multirow{3}{*}{Frame number} 
& 	Average	&	70	&	34 &	73\\
& 	Maximum	&	126	&	58 &	101\\
& 	Minimum	&	24	&	11 & 30\\
\hline

\multirow{4}{*}{\begin{tabular}[c]{@{}l@{}} Video duration\\ (\textit{s})\end{tabular}}  
& 	Average	&	0.35	&	0.34 &	0.36\\
& 	Maximum	&	0.63	&	0.58 &	0.51\\
& 	Minimum	&	0.12	&	0.11 &  0.15\\
\hline

\multirow{3}{*}{Expression} 
& 	Negative	&	88	&	70 &	92\\
& 	Positive	&	32	&	51 &	26\\
& 	Surprise	&	25	&	43 & 15\\
\hline

\multirow{4}{*}{\begin{tabular}[c]{@{}l@{}} Ground-truth\\annotations\end{tabular}}   
& 	Onset index	&	\cmark	&	\cmark	&	\cmark\\
& 	Offset index	&	\cmark	&	\cmark	&	\cmark\\
& 	Apex index	&	\cmark	&	\xmark	&	\cmark\\
\hline

\end{tabular}
\end{center}
\end{table}

\setlength{\tabcolsep}{4pt}
\begin{table*}[t!]
\begin{center}
\caption{Comparison of micro-expression recognition performance in terms of Accuracy (\textit{Acc}), F1-score, Unweighted F1-score (\textit{UF1}) and Unweighted Average Recall (\textit{UAR}) on the composite (\textit{Full}), CASME II, SMIC and SAMM databases}
\label{table:result}
\begin{tabular}{clcccccccccc}

\hline
\multirow{2}{*}{No.}
&\multirow{2}{*}{Methods}
& \multicolumn{4}{c}{Full}
& \multicolumn{2}{c}{SMIC} 
& \multicolumn{2}{c}{CASME II}
& \multicolumn{2}{c}{SAMM}
\\
\cline{3-12}
&  
& Acc
& F1-score
& UF1
& UAR 
& UF1
& UAR 
& UF1
& UAR 
& UF1
& UAR \\
\hline

1
& LBP-TOP baseline 
& - 
& -
& 0.5882
& 0.5785
& 0.2000
& 0.5280
& 0.7026
& 0.7429
& 0.3954
& 0.4102\\

2
& Bi-WOOF~\cite{liong2018less}
& 0.6833
& 0.6304
& 0.6296
& 0.6227
& 0.5727
& 0.5829
& 0.7805
& 0.8026
& 0.5211
& 0.5139 \\

3
& AlexNet~\cite{krizhevsky2012imagenet}
& 0.7308
& 0.6959
& 0.6933
& 0.7154
& 0.6201
& 0.6373
& 0.7994
& 0.8312	
& 0.6104
& 0.6642 \\

4
& SqueezeNet~\cite{iandola2016squeezenet}
& 0.6380
& 0.5964
& 0.5930	
& 0.6166
& 0.5381
& 0.5603
& 0.6894
& 0.7278
& 0.5039
& 0.5362\\

5
& GoogLeNet~\cite{szegedy2015going}
& 0.6335
& 0.5698
& 0.5573
& 0.6049	
& 0.5123	
& 0.5511	
& 0.5989
& 0.6414
& 0.5124
& 0.5992
\\

6
& VGG16~\cite{simonyan2014very}
& 0.6833
& 0.6439
& 0.6425
& 0.6516
& 0.5800
& 0.5964
& 0.8166
& 0.8202	
& 0.4870
& 0.4793
\\

7
& OFF-ApexNet~\cite{liong2018off}
& 0.7460	
& 0.7104
& 0.7196
& 0.7096
& 0.6817
& 0.6695
& 0.8764
& 0.8681
& 0.5409
& 0.5392 \\

8
& \textbf{STSTNet}
& \textbf{0.7692}	
& \textbf{0.7389}
& \textbf{0.7353}
& \textbf{0.7605}
& \textbf{0.6801}
& \textbf{0.7013}
& \textbf{0.8382}
& \textbf{0.8686}
& \textbf{0.6588}	
& \textbf{0.6810}\\

\hline 

\end{tabular}
\end{center}
\end{table*}

\subsection{Performance Metric}
As the emotion classes are still imbalanced after merging (250 Negative, 109 Positive, 83 Surprise), 
two balanced metrics are used to reduce potential bias: Unweighted F1-score (UF1) 
and Unweighted Average Recall (UAR). Intuitively, both metrics are the averaged per-class computation of their respective original metrics: 
%
\begin{equation}\label{eq:acc}
\text{Accuracy} := \frac{\sum_{\alpha=1}^M \sum_{\beta=1}^k {TP_\alpha^{\beta}}}{\sum_{\alpha=1}^M \sum_{\beta=1}^k{TP_\alpha^k}+\sum_{\alpha=1}^M\sum_{\beta=1}^k{FP_\alpha^k}}
\end{equation}
\begin{equation}\label{eq:f-measure}
\text{F1-score} := \frac{2 \times \text{Precision} \times \text{Recall}}{\text{Precision} + \text{Recall}}
\end{equation}
\begin{equation}\label{eq:recall_f1}
\text{Recall} := {\sum_{\alpha=1}^M \frac{\sum_{\beta=1}^k TP_\alpha^\beta}{M \times \sum_{\beta=1}^k TP_\alpha^\beta+ \sum_{\beta=1}^k FN_\alpha^\beta}}
\end{equation}
\begin{equation}\label{eq:precision_f1}
\text{Precision} := {\sum_{\alpha=1}^M \frac{\sum_{\beta=1}^k TP_\alpha^\beta}{M \times \sum_{\beta=1}^k TP_\alpha^\beta+ \sum_{\beta=1}^k FP_\alpha^\beta}}
\end{equation}
where $M$ is the number of classes; TP, FN and FP are the true positive, false negative and false positive, respectively.

The new balanced metrics are expressed as follows:
\begin{equation}\label{eq:Uf-measure}
\text{UF1} := 2 \times \frac{\sum_{\alpha=1}^M \frac{\text{Precision}_\alpha \times \text{Recall}_\alpha}{\text{Precision}_\alpha + \text{Recall}_\alpha}}{M} 
\end{equation}
\begin{equation}\label{eq:precision_balanced}
\text{Precision}_\alpha := {\frac{\sum_{\beta=1}^k TP_\alpha^\beta}{\sum_{\beta=1}^k TP_\alpha^\beta+ \sum_{\beta=1}^k FP_\alpha^\beta}}
\end{equation}
\begin{equation}\label{eq:recall_balanced}
\text{Recall}_\alpha := {\frac{\sum_{\beta=1}^k TP_\alpha^\beta}{\sum_{\beta=1}^k  TP_\alpha^\beta+ \sum_{\beta=1}^k FN_\alpha^\beta}}
\end{equation}
\begin{equation}\label{eq:UAR}
\text{UAR} := \frac{1}{M}\sum_{\alpha=1}^M \frac{\sum_{\beta=1}^k TP_\alpha^\beta}{M \times \sum_{\beta=1}^k TP_\alpha^\beta+ \sum_{\beta=1}^k FN_\alpha^\beta}
\end{equation}

All the results presented are \ice{evaluated based on} leave-one-subject-out (LOSO) cross validation protocol. 

\section{Results and Discussion} 

We implemented a number of benchmark methods, i.e. LBP-TOP~\cite{zhao2007}) baseline, state-of-the-art methods of Bi-WOOF~\cite{liong2018less} and OFF-ApexNet~\cite{liong2018off}), and some popular deep learning architectures (AlexNet~\cite{krizhevsky2012imagenet}, Squeezenet~\cite{iandola2016squeezenet}, GoogLeNet~\cite{szegedy2015going}, VGG16~\cite{simonyan2014very}) in their original form, to compare against the proposed method. Table~\ref{table:result} reports our results.
\textit{Full} refers to the composite database.
\ice{Table~\ref{table:result} shows that} the proposed STSTNet outperforms other methods in all scenarios, except for OFF-ApexNet on CASME II. More importantly, it achieved the best UF1 (0.7353) and UAR (0.7605) on the full composite database.
Overall, the STSTNet approach produced an average improvement of approximately 15\%, 48\%,  14\%, 26\% over the LBP-TOP baseline on the composite, SMIC, CASME II and SAMM databases respectively.


\ice{The confusion matrix} in Table~\ref{table:confusion} \ice{shows} that the proposed method is capable at distinguishing the negative emotion, which is partly attributed to the fact that
more than half of all videos belong to the Negative class.
\christy{In SAMM, the STSTNet also performed very well on the Negative class ($\sim$90\%); the class imbalanced problem is even more severe in the case of SAMM.} 
As expected, the high frame rate capture of the CASME II samples provides more precise apex frames, which in turn, leads to more accurate optical flow computation that better characterizes the motion changes.
On the contrary, STSTNet exhibits lower recognition performance on the SMIC as compared to CASME II due two possible reasons. First, the addition of the spotting step (which has a reported MAE of $\sim$13 frames~\cite{liong2015spotting}) may introduce potential errors due to the inaccurately spotted apex frames. 
Besides, SMIC videos are captured at a lower frame rate (100 fps), and are affected by various background noises such as the shadows, highlights, illumination, flickering lights due to the database elicitation setup.

\begin{table}[thpb]
	\begin{center}
    \caption{The confusion matrix of STSTNet on \textit{Full}, SMIC, CASME II and SAMM databases (measured by recognition rate \%)}
    \label{table:confusion}
    \begin{subtable}[l]{.48\linewidth}
    \centering
        \caption{Full}
        \begin{tabular}{cccc}
        \hline
         & Neg	 
         & Pos 
         & Sur \\
        \hline
        Neg  
        & \textbf{87.60}
        & 8.80
        & 3.60\\
        Pos
        & 36.70	
        & \textbf{56.88}
        & 6.42\\
        Sur
        & 25.30
        & 3.61
        & \textbf{71.08}\\
        \hline
        \end{tabular}
    \end{subtable}%
    \begin{subtable}{.48\linewidth}
    \centering 
        \caption{SMIC}
        \begin{tabular}{cccc}
        \hline
         & Neg	 
         & Pos 
         & Sur \\
        \hline
        Neg  
        & \textbf{77.14}
        & 14.29
        & 8.57\\
        Pos
        & 33.33	
        & \textbf{58.82}
        & 7.84\\
        Sur
        & 32.56
        & 2.33
        & \textbf{65.12}\\
        \hline
        \end{tabular}
    \end{subtable} 
    
    \bigskip
    
    \begin{subtable}[l]{.48\linewidth}
    \centering
        \caption{CASME II}
        \begin{tabular}{cccc}
        \hline
         & Neg	 
         & Pos 
         & Sur \\
         
        \hline
        
        Neg  
        & \textbf{94.32}
        & 5.68
        & 0\\

        Pos
        & 37.50	
        & \textbf{59.38}
        & 3.13\\

        Sur
        & 8
        & 0
        & \textbf{92}\\
        \hline
        \end{tabular}
    \end{subtable}%
    \begin{subtable}{.48\linewidth}
    \centering  
        \caption{SAMM}
        \begin{tabular}{cccc}
        \hline
         & Neg	 
         & Pos 
         & Sur \\
         
        \hline
        
        Neg  
        & \textbf{89.13	}
        & 7.61
        & 3.26\\

        Pos
        & 42.31	
        & \textbf{50.00}
        & 7.69\\

        Sur
        & 33.33
        & 13.33
        & \textbf{53.33}\\
        \hline
        \end{tabular}
    \end{subtable}

    \end{center}
\end{table}

Table~\ref{table:properties} summarizes some key properties of all competing neural networks mentioned in Table~\ref{table:result}, \ice{including}: 1) Depth - the largest number of sequential convolutional or fully connected layers in an end-to-end neural network; 2) Learnable parameters - the number of weights and biases in the network; 3) Image input size - \ice{input image resolution}; 4) Fold execution time - The total training and testing time for a single fold of LOSO cross validation evaluation. 
The STSTNet has the least network depth (2), learnable parameters size (1670 weights and biases) and \christy{a relatively low computational time} ($\sim~5.7 s$ to train the model and infer test data).
\christy{Our network models are implemented in MATLAB, and code is publicly available\footnote{\href{https://github.com/christy1206/STSTNet}{https://github.com/christy1206/STSTNet}}} for non-commercial, or research use.

\begin{table}[t] 
\begin{center}
\caption{Key properties of the competing neural networks
}
\label{table:properties}
\begin{tabular}{lcccc}
\hline

\multirow{2}{*}{Network}
& \multirow{2}{*}{Depth}
& \multirow{2}{*}{\begin{tabular}[c]{@{}l@{}}Parameter\\  (Million)\end{tabular}}
& \multirow{2}{*}{Image Input Size}
& \multirow{2}{*}{\begin{tabular}[c]{@{}l@{}}Execution  \\  Time (\textit{s})\end{tabular}}\\\\
\hline 

\textbf{STSTNet}
& \textbf{2}
& \textbf{0.00167}
& 28 $\times$ 28 $\times$ 3
& 5.7366
\\

OFF-ApexNet~\cite{liong2018off}
& 5
& 2.77
& 28 $\times$ 28 $\times$ 2
& 5.5632
\\

AlexNet~\cite{krizhevsky2012imagenet}
& 8
& 61
& 227 $\times$ 227 $\times$ 3
& 12.9007
\\

SqueezeNet~\cite{iandola2016squeezenet}
& 18
& 1.24
& 227 $\times$ 227 $\times$ 3
& 14.3704
\\

GoogLeNet~\cite{szegedy2015going}
& 22
& 7
& 224 $\times$ 224 $\times$ 3
& 29.3022
\\

VGG16~\cite{simonyan2014very}
& 16
& 138
& 224 $\times$ 224 $\times$ 3
& 95.4436
\\
\hline
\end{tabular}
\end{center}
\end{table}

\section{Conclusion}
As a submission entry to the 2nd Micro-Expression  Grand Challenge (MEGC), this paper presents a novel shallow triple stream three-dimensional CNN (STSTNet) to learn optical flow guided features for ME recognition.
A compact and discriminative feature representation is learned from an input cube consisting of three optical flow images (i.e., horizontal optical flow, vertical optical flow and optical strain).
Overall, the proposed STSTNet approach \ice{has} demonstrated promising recognition results on a newly merged composite ME database consisting of three spontaneous ME databases, yielding a UF1 of 0.7353 and UAR of 0.7605 which surpassed recent  state-of-the-art methods. 
In future, the apex spotting technique requires further improvement to extract more accurate apex frames for ME recognition. A number of recent works have begun to exploit the benefits of using the apex frame for recognition \cite{liong2018less}. Furthermore, other measures of the optical flow field such as the magnitude and orientation can also be considered as input \ice{to} the neural network.

\section*{Acknowledgements}

This work was funded in part by Ministry of Science and Technology (MOST) (Grant Number: MOST107-2218-E-035-006-), MOHE Grant FRGS/1/2016/ICT02/MMU/02/2 Malaysia and Shanghai 'The Belt and Road' Young Scholar Exchange Grant (17510740100).

\bibliographystyle{ieee}
\bibliography{refs}
\end{document}